\documentclass[10pt,twocolumn,letterpaper]{article}
\usepackage[pagenumbers]{cvpr} 
\usepackage{graphicx}
\usepackage{subcaption}
\usepackage{booktabs}
\usepackage{multirow}
\usepackage{array}
\usepackage{algpseudocode}

\definecolor{cvprblue}{rgb}{0.21,0.49,0.74}
\definecolor{darkred}{HTML}{8B0000}
\usepackage[pagebackref,breaklinks,colorlinks,allcolors=cvprblue]{hyperref}

\def\paperID{11} 
\def\confName{ICCV}
\def\confYear{2025}

\title{ Are you In or Out (of gallery)? Wisdom from the Same-Identity Crowd}

\author{Aman Bhatta$^1$ \quad\quad Maria Dhakal$^1$ \quad\quad  Michael C. King$^2$ \quad\quad Kevin W. Bowyer$^1$${\thanks{Dr. Bowyer is a member of the FaceTec (\url{facetec.com}) Advisory Board.  Results in this paper do not necessarily relate to FaceTec products.}}$ \\\\
$^1$University of Notre Dame\\
$^2$Florida Insitute of Technology\\
{{\tt\small \{abhatta,kwb\}@nd.edu,michaelking@fit.edu}}}

\begin{document}
\maketitle
\begin{abstract}

A central problem in one-to-many facial identification is that the person in the probe image may or may not have enrolled image(s) in the gallery; that is, may be In-gallery or Out-of-gallery.  
Past approaches to detect when a rank-one result is Out-of-gallery have mostly focused on finding a suitable threshold on the similarity score.  We take a new approach, using the additional enrolled images of the identity with the rank-one result to predict if the rank-one result is In-gallery / Out-of-gallery. Given a gallery of identities and images, we generate In-gallery and Out-of-gallery training data by extracting the ranks of additional enrolled images corresponding to the rank-one identity. We then train a classifier to utilize this feature vector to predict whether a rank-one result is In-gallery or Out-of-gallery. Using two different datasets and four different matchers, we present experimental results showing that our approach is viable for mugshot quality probe images, and also, importantly, for probes degraded by blur, reduced resolution, atmospheric turbulence and sunglasses.  We also analyze results across demographic groups, and show that In-gallery / Out-of-gallery classification accuracy is similar across demographics.  Our approach has the potential to provide an objective estimate of whether a one-to-many facial identification is Out-of-gallery, and thereby to reduce false positive identifications, wrongful arrests, and wasted investigative time. Interestingly, comparing the results of older deep CNN-based face matchers with newer ones suggests that the effectiveness of our Out-of-gallery detection approach emerges only with matchers trained using advanced margin-based loss functions.
\vspace{-1.5em} 
\end{abstract}

\section{Introduction}
Concern has arisen over the use of one-to-many facial identification, particularly in law enforcement, where false arrests occur due to the rank-one identified individual not being in the gallery \cite{report_aclu_2018,hoggins_telegraph_2019}. Most academic research focuses on closed-set identification, assuming all probe images exist in the gallery, whereas real-world deployments must handle open-set conditions where probes may have no corresponding image(s) in the gallery. 


\begin{figure}[t]
    \centering
    \includegraphics[width=\linewidth, height=8cm]{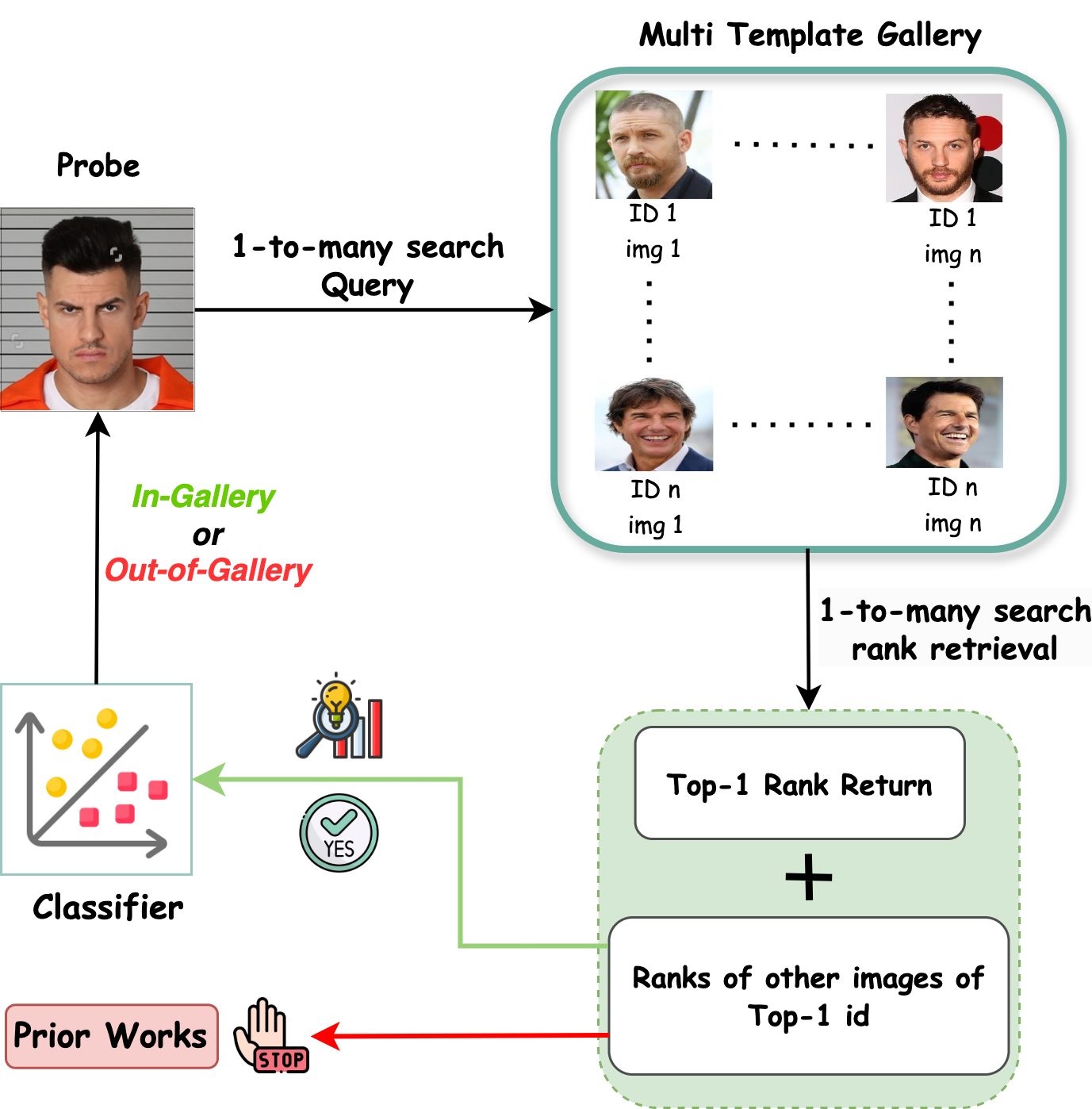}
    \caption{
{\bf Classifying In-Gallery vs. Out-of-Gallery Samples Using Rank Patterns}. The ranks of additional images for a rank-one identity in a 1-to-many search reveal whether a sample is in-gallery or out-of-gallery.
    }
    \vspace{-1.75em}
    \label{fig:makeup_dist}
\end{figure}

A naive approach to distinguishing ``In-gallery” vs. ``Out-of-gallery” samples in 1-to-many search is to apply a threshold on similarity scores. To address this limitation, researchers have proposed open-set classification protocols, often evaluated on the LFW dataset \cite{gunther_cvprw_2017,liao_ijcb_2014}. These protocols typically use the maximum or average similarity score from multiple enrolled images and apply thresholding. EVT-based approaches refine open-set recognition by modeling uncertainty, while distance-based methods, thresholding techniques, and machine learning adaptations aim to improve decision boundaries. Deep learning strategies further enhance open-set handling by introducing specialized architectures and loss functions \cite{bendale_cvpr_2015b}. However, these methods often rely on predefined thresholds, statistical assumptions, or distance heuristics that may not generalize well across diverse data distributions. Additionally, deep learning-based solutions, while effective, frequently require extensive retraining or architectural modifications, limiting their scalability in real-world deployments.

In this work, we aim to learn a classifier that determines whether a rank query vector from a probe search corresponds to an In-gallery or Out-of-gallery sample. 
Modern face recognition systems typically use multiple images per identity, as evidenced by NIST's evaluation of 30.2 million photos from 14.4 million identities \cite{grother_nistir_2019b}. Given multi-image enrollment per identity, a 1-to-many probe search returns a rank-one identity along with the ranks of other images of that identity. Previous works have largely ignored the information available from the non-rank-one images of the rank-one identity.
However, we exploit the results of the non-rank-one images as the foundation of our approach to classifying the rank-one identity as In-gallery / Out-of-gallery. Traditionally, a threshold ($\alpha$) is used to accept or reject the rank-one identity as an ``In-gallery” or ``Out-of-gallery” sample, but it requires careful tuning and is often unreliable in real-world scenarios. Additionally, open-set face identification must balance \textit{known unknowns}—probes from identities seen during training but absent in the gallery—and \textit{unknown unknowns}—probes from identities never encountered during training or enrollment.

Instead of relying on these distinctions, our approach simplifies the problem by performing two searches for each probe. In the first search scenario (\textit{In-gallery}), additional images of the probe are included in the gallery, and non-rank-one rankings are retrieved to generate query vectors labeled as “In-gallery.” In the second search scenario (\textit{Out-of-gallery}), the probe’s additional images are excluded from the gallery, and non-rank-one rankings of the identity that appears as rank-one are retrieved to generate query vectors labeled as “Out-of-gallery.” This approach balances the dataset, ensuring stable learning and improved generalization by equally representing both scenarios. {\it Notably, in the \textit{In-gallery} search scenario, the rank-one identity is typically correct due to the robustness of modern face embedding networks. However, older face embedding networks suffer from this issue, leading to degraded performance in downstream classification, as discussed in Section \ref{results}.} Furthermore, our approach does not rely on statistical distribution assumptions and remains agnostic to the frequency of \textit{In-gallery} and \textit{Out-of-gallery} scenarios during inference, enhancing its adaptability to real-world conditions.

\noindent The core contributions of this paper include:

\begin{itemize}  
    \item We introduce a novel approach for classifying facial identification results as In-gallery / Out-of-gallery, leveraging the rank information of images beyond the rank-one image of the identity—an aspect not explored in prior work.
    \item We present the first analysis of In-gallery / Out-of-gallery prediction accuracy for probe images with real-world degradations such as blur, decreased resolution, atmospheric turbulence, and occlusions such as sunglasses. Figure 6 shows that the better face matchers produce rank feature vectors that still allow remarkable accuracy even for degraded quality probes. 
    \item We introduce demographic-aware probe-gallery sampling for In-gallery vs. Out-of-gallery prediction, motivated by real-world false arrests, which predominantly occur within a demographic group rather than across demographics. Our analysis shows that In-gallery / Out-of-gallery prediction accuracy varies only slightly across demographics, with female probes having slightly lower accuracy in some conditions.  (See Figure \ref{fig:MORPH}.) 
\end{itemize}

This paper is organized as follows. Section \ref{litreview} provides a brief literature review. Section \ref{face} details the dataset, sampling strategy, image processing pipeline, and face embedding networks used. Section \ref{prelim} outlines the motivations and observations that lay the foundation for the core contributions. Section \ref{method} presents train-test data curation, training procedures, and evaluation protocols. Section \ref{results} reports the experimental results, followed by Section \ref{discussions}, which provides an in-depth discussion of the proposed approach. Finally, Section \ref{conclusion} concludes the paper.
\vspace{-0.5em}
\section{Literature Review}\label{litreview}

A key challenge in one-to-many search is its open-set nature, where a probe may not have a corresponding match in the gallery. 
A search without a corresponding match in the gallery inevitably results in a false-positive identification, potentially leading to consequences such as wrongful arrests—an issue that has drawn scrutiny from both media and academic communities \cite{report_aclu_2018,hoggins_telegraph_2019,general_cnn_2021}. Most open-set biometric systems employ a {\it threshold-based} approach, where a probe is labeled as ``not in gallery" if its highest match score falls below a threshold \cite{grother_nist_2019_2}. This threshold must balance the False Positive Identification Rate (FPIR), where an Out-of-gallery probe is incorrectly matched to an enrolled identity, and the False Negative Identification Rate (FNIR), where an In-gallery probe fails to match. However, using a fixed rejection threshold is problematic, as the notion of ``unknown" evolves with increasing enrollments and varying data distributions, making a one-size-fits-all threshold ineffective \cite{gunther_cvprw_2017}. Lowering the threshold reduces FNIR but increases FPIR, and vice versa. To address these limitations, methods such as Extreme Value Theory (EVT) and multi-class unknown modeling have significantly advanced open-set recognition in biometrics \cite{scheirer_tpami_2013,rattani_tifs_2015}. Scheirer et al. introduced meta-recognition techniques and later proposed W-SVM and CAP models, leveraging EVT-based probability estimates to detect unknown classes \cite{scheirer_tpami_2011,scheirer_tpami_2013}. This led to the development of the Extreme Value Machine (EVM), which establishes robust open-set decision boundaries \cite{scheirer_tpami_2014}. In parallel, Bendale et al. adapted deep networks for open-set recognition by calibrating final-layer activations via EVT, enabling reliable rejection of unseen classes \cite{bendale_cvpr_2015a,bendale_cvpr_2015b}. These approaches have improved biometric recognition across modalities, including face recognition and fingerprint spoof detection, as highlighted in \cite{geng_tpami_2021}. Other recent works include \cite{vareto_ivc_2024,cruz_eccv_2024}. 

Traditional face recognition, including EVT-based open-set protocols, often assume a single score per subject, typically the max similarity score between the probe and enrolled images or the average cosine similarity score \cite{gunther_cvprw_2017}. Despite its limitations, this single-score approach remains prevalent, particularly in large-scale or unconstrained settings \cite{grother_nistir_2019b}. Instead of relying on simple averaging, EVT methods can be improved by leveraging multiple images (or video frames) to represent each subject \cite{yang_cvpr_2017}. Basic approaches average deep embeddings, while more advanced methods employ attention mechanisms or learned fusion techniques \cite{yang_cvpr_2017,bodla_wacv_2017,kim_neurips_2022,shi_cvpr_2019,jawade_neurips_2025}. Studies consistently show that multi-image enrollment enhances one-to-many identification by reducing false matches and increasing robustness to covariates like pose and aging \cite{Klare2015}. However, fusing results from multiple gallery images and deriving a unified score for threshold-based matching still suffers from the inherent limitations of score thresholding, even if it improves accuracy of closed-set face recognition.  

In this work, we leverage the rank patterns of multiple images for the rank-one identity to determine whether a match is an ``In-gallery" or ``Out-of-gallery". This mitigates the limitations of threshold-based methods by eliminating the need for manual threshold tuning and addresses the weaknesses of statistical EVT-based approaches, which depend on distributional assumptions.

\section{Face Representation Learning}\label{face}

\noindent{\bf Datasets.} We use MORPH and the ND-Male-Female Accuracy Dataset (MFAD). These datasets include demographic labels, enabling an analysis of the proposed approach across demographic groups. MORPH \cite{morph_site_github, ricanek_afgr_2006} is widely used in face aging research and demographic studies~\cite{Drozdowski_TTS_2020}. It contains 35,276 images of 8,835 Caucasian males, 10,941 images of 2,798 Caucasian females, 56,245 images of 8,839 African-American males, and 24,855 images of 5,928 African-American females, with an average of 4–6.4 images per identity. MFAD \cite{tian_fg_2025} results from a different controlled image acquisition, has sunglasses-wearing images for probes and has IRB-approved consent for research use. The dataset includes 6,669 images of 575 Caucasian females and 8,419 images of 687 Caucasian males. 

Past work has used academic benchmarks like LFW to define open-set protocols for face recognition \cite{gunther_cvprw_2017}, but MORPH better reflects real-world surveillance galleries. Moreover, prior open-set recognition research has not examined its impact across demographic groups. With identity and demographic labels available in MORPH, we can analyze performance across demographics. Since ND-MFAD is collected in a different setting, evaluating our framework on ND-MFAD helps assess its generalizability and ensures our findings are not artifacts specific to MORPH.
\begin{figure}[!htbp]
\centering
  \begin{subfigure}[b]{0.95\linewidth}
    \centering
      \begin{subfigure}[b]{0.24\linewidth}
        \centering
          \includegraphics[width=1\linewidth]{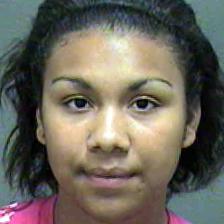}
          \caption{Original}
      \end{subfigure}
      \hfill
      \begin{subfigure}[b]{0.24\linewidth}
        \centering
          \includegraphics[width=1\linewidth]{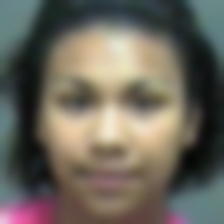}
          \caption{Blur}
      \end{subfigure}
      \hfill
      \begin{subfigure}[b]{0.24\linewidth}
        \centering
          \includegraphics[width=1\linewidth]{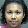}
          \caption{Downsample}
      \end{subfigure}
      \hfill
      \begin{subfigure}[b]{0.24\linewidth}
        \centering
          \includegraphics[width=1\linewidth]{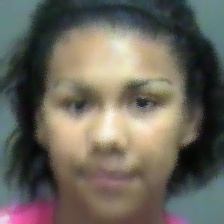}
          \caption{At. Turb}
      \end{subfigure}
  \end{subfigure}

  \begin{subfigure}[b]{0.95\linewidth}
    \centering
      \begin{subfigure}[b]{0.24\linewidth}
        \centering
          \includegraphics[width=1\linewidth]{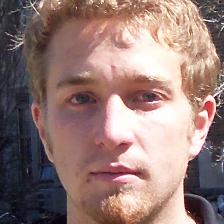}
          \caption{Original}
      \end{subfigure}
      \hfill
      \begin{subfigure}[b]{0.24\linewidth}
        \centering
          \includegraphics[width=1\linewidth]{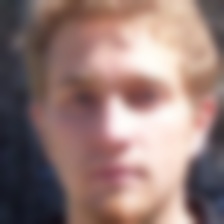}
          \caption{Blur}
      \end{subfigure}
      \hfill
      \begin{subfigure}[b]{0.24\linewidth}
        \centering
          \includegraphics[width=1\linewidth]{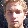}
          \caption{Downsample}
      \end{subfigure}
      \hfill
      \begin{subfigure}[b]{0.24\linewidth}
        \centering
          \includegraphics[width=1\linewidth]{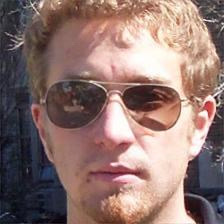}
          \caption{Sunglasses}
      \end{subfigure}
  \end{subfigure}
  \vspace{-0.75em} 
  \caption{Sample MORPH (a to d) and MFAD (e to h) images.}
  \label{fig:sample_images}
  \vspace{-1em}
\end{figure}

\noindent{\bf Probe and Gallery Image Processing.} To ensure the robustness of the method, we evaluate probes under various conditions, including blur, downsampling, atmospheric turbulence, and sunglasses, in addition to the original probes. While deep CNN matchers are highly effective, prior works~\cite{pangelinan_tts_2024} have shown that state-of-the-art matchers fail under blur and downsampling conditions. Following~\cite{pangelinan_tts_2024}, we apply Gaussian smoothing with $\sigma = 5$ to generate blurred probe images. Low-resolution face images are also common in surveillance settings. Following~\cite{pangelinan_tts_2024}, we use bicubic interpolation to downsample the original 224×224 probe images to 28×28. For ND-MFAD, we apply the same blurring and downsampling procedures and include probe images with sunglasses, as provided in~\cite{tian_fg_2025}. To simulate turbulence conditions, we use the turbulence simulator proposed by Mao et al. \cite{mao_iccv_2021}. The effect of various turbulence settings on deep face embedding networks has been studied by \cite{robbins_cvprw_2022}. Based on their recommendations, we adopt a D/$r_o$ setting of 1.5, which corresponds to a turbulence range from 0 (no turbulence) to 5 (extreme turbulence). Beyond a D/$r_o$ ratio of 1.5, the face recognition models start to treat the distortions as salient features, which negatively impacts the face recognition performance. {\it Note that only probe images are altered in experiments shown in the main paper.} Sample images are shown in Figure~\ref{fig:sample_images}.

\noindent{\bf Face Embedding Networks.} We evaluate four face embedding networks, from earlier models to current state-of-the-art, including ViT-based architectures. First, we use FaceNet~\cite{schroff_cvpr_2015}, a pre-margin-based loss model with an Inception ResNetV1 backbone~\cite{szegedy_aaai_2017} and weights from~\cite{facenet_github_2016}, to assess performance with a less accurate embedding model. Second, we evaluate an ArcFace-based model~\cite{deng_cvpr_2019} trained on a ResNet-101 backbone with weights from~\cite{insightface_github_2016}. Third, we include AdaFace, an adaptive margin model designed for low-quality images, trained on a ResNet-101 backbone with modified loss and augmentations, using weights from~\cite{insightface_github_2016}. Finally, we assess TransFace~\cite{dan_iccv_2023}, a ViT-based model trained  with ArcFace loss, incorporating patch-level augmentation and hard sample mining, with weights from~\cite{transface_github_2023}. This selection ensures evaluation of \textit{In-gallery} vs. \textit{Out-of-gallery} classification across diverse embedding approaches. {\it For each face embedding network, we use open-source weights that achieve the highest average accuracy on FR benchmarks for that model.} Each network takes an aligned 112×112 face image as input and outputs a 512-d feature vector, matched using cosine similarity. Face detection and alignment are performed using img2pose~\cite{albiero_cvpr_2021}.

\section{Motivations and Preliminary}\label{prelim}
\noindent{\bf Motivations.} Any probe used in a one-to-many search is guaranteed to return a gallery image as its rank-one match—the image with the highest similarity score to the probe. However, determining whether this rank-one match is an \textit{In-gallery} or \textit{Out-of-gallery} sample based solely on the similarity score is unreliable. In real-world systems, the enrolled identity associated with the rank-one image typically has multiple images in the gallery \cite{grother_nistir_2019b}. If the rank-one identity has additional images in the gallery, the ranks of these beyond-rank-one images may provide valuable information for determining whether the rank-one match is an \textit{In-gallery} or \textit{Out-of-gallery} sample. We hypothesize that when the rank-one identity is a \textit{Out-of-gallery} sample, its selection may be driven by incidental similarities such as facial expression, beard, or illumination artifacts. Since the additional enrolled images of that identity may not share these incidental features, their ranks should be significantly displaced from rank-one. Conversely, for an \textit{In-gallery} rank-one identity, the additional enrolled images should typically rank close to rank one, reflecting the robustness of modern deep learning face embedding networks. 

\noindent{\bf Demographic-Aware Probe and Gallery Sampling.}\label{cross-demo} Before presenting the motivation, we first outline the sampling process for probes and galleries in our experiments. Given a gallery where each identity has multiple images, the most recent image is selected as the probe, while the remaining images are enrolled in the gallery, mirroring real-world scenarios. A potential concern is the occurrence of rank-one matches across demographic groups. While cross-demographic, non-mated rank-one matches can occur, prior work \cite{pangelinan_tts_2024} has shown them to be rare. Moreover, documented cases of false arrests consistently involve individuals of the same demographic group as the probe used in the search \cite{report_aclu_2018,hoggins_telegraph_2019}. Accordingly, in our experiments, {\it all gallery images used for each probe’s 1-to-many search are strictly sampled from the same demographic group.}

\begin{figure*}[!htbp]
  \begin{subfigure}[b]{1\linewidth}
    \centering
      \begin{subfigure}[b]{0.24\linewidth}
        \centering
          \includegraphics[width=1\linewidth]{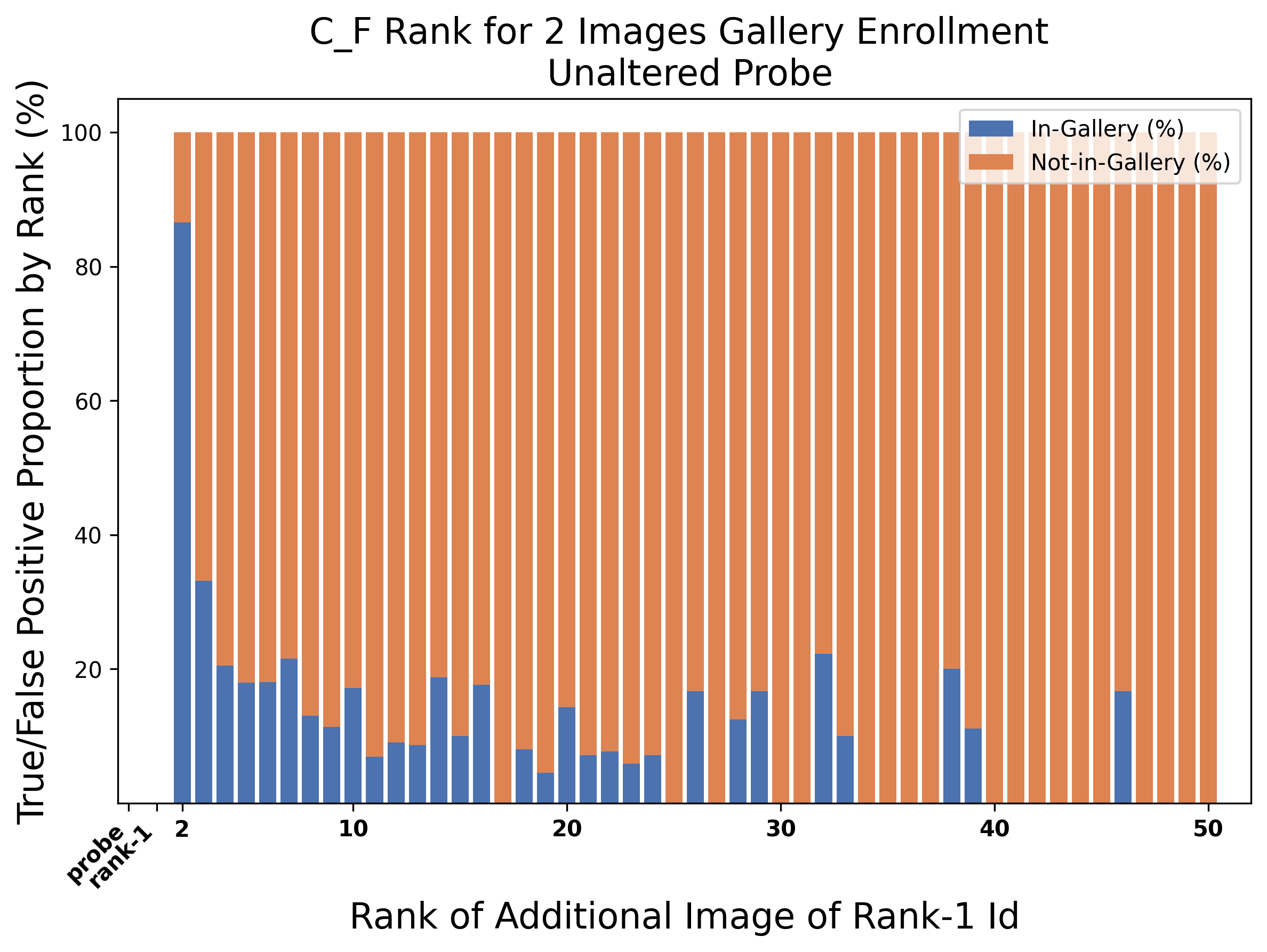}
      \end{subfigure}
      \hfill
      \begin{subfigure}[b]{0.24\linewidth}
        \centering
          \includegraphics[width=1\linewidth]{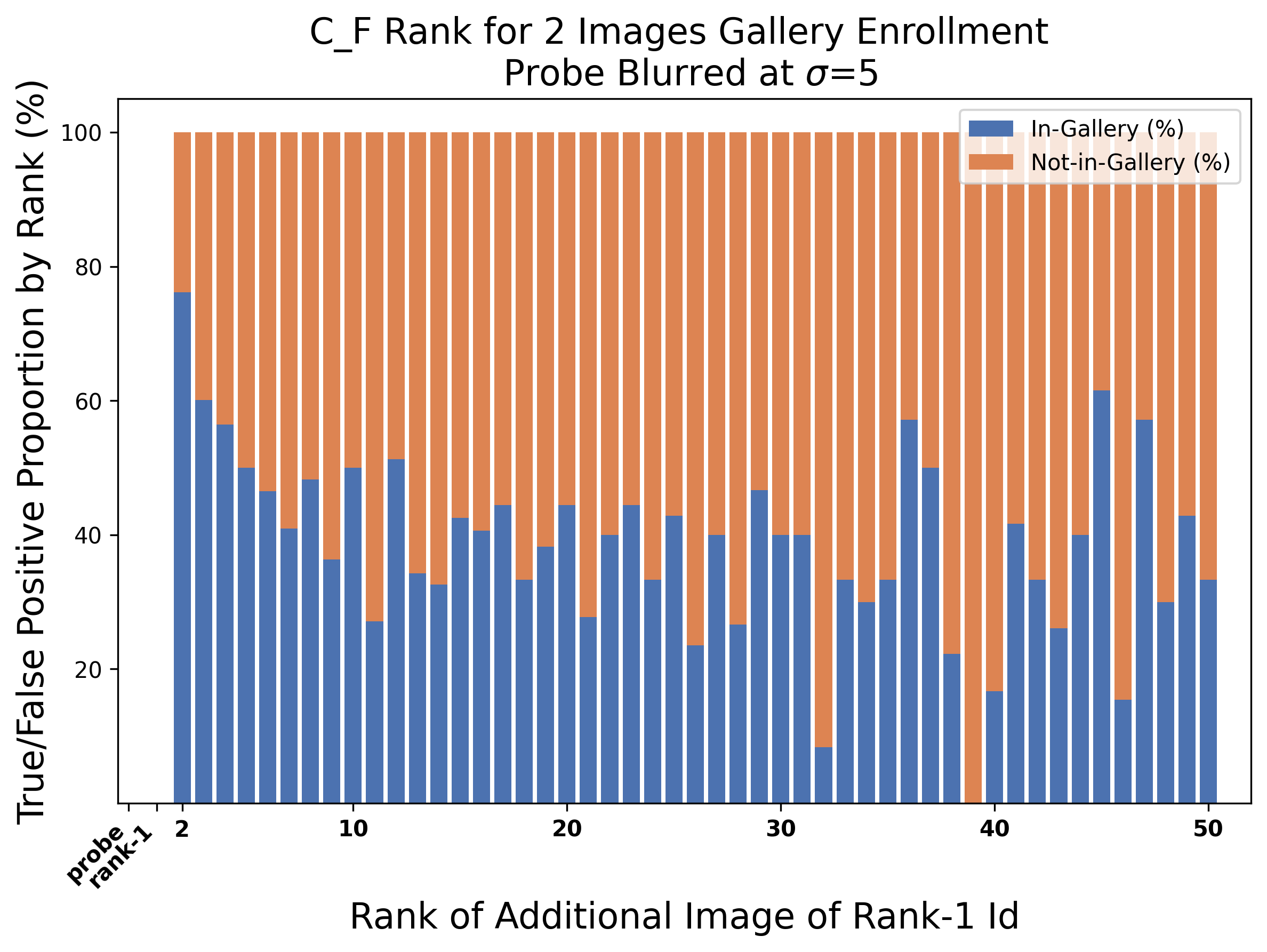}
      \end{subfigure}
      \hfill
      \begin{subfigure}[b]{0.24\linewidth}
        \centering
          \includegraphics[width=1\linewidth]{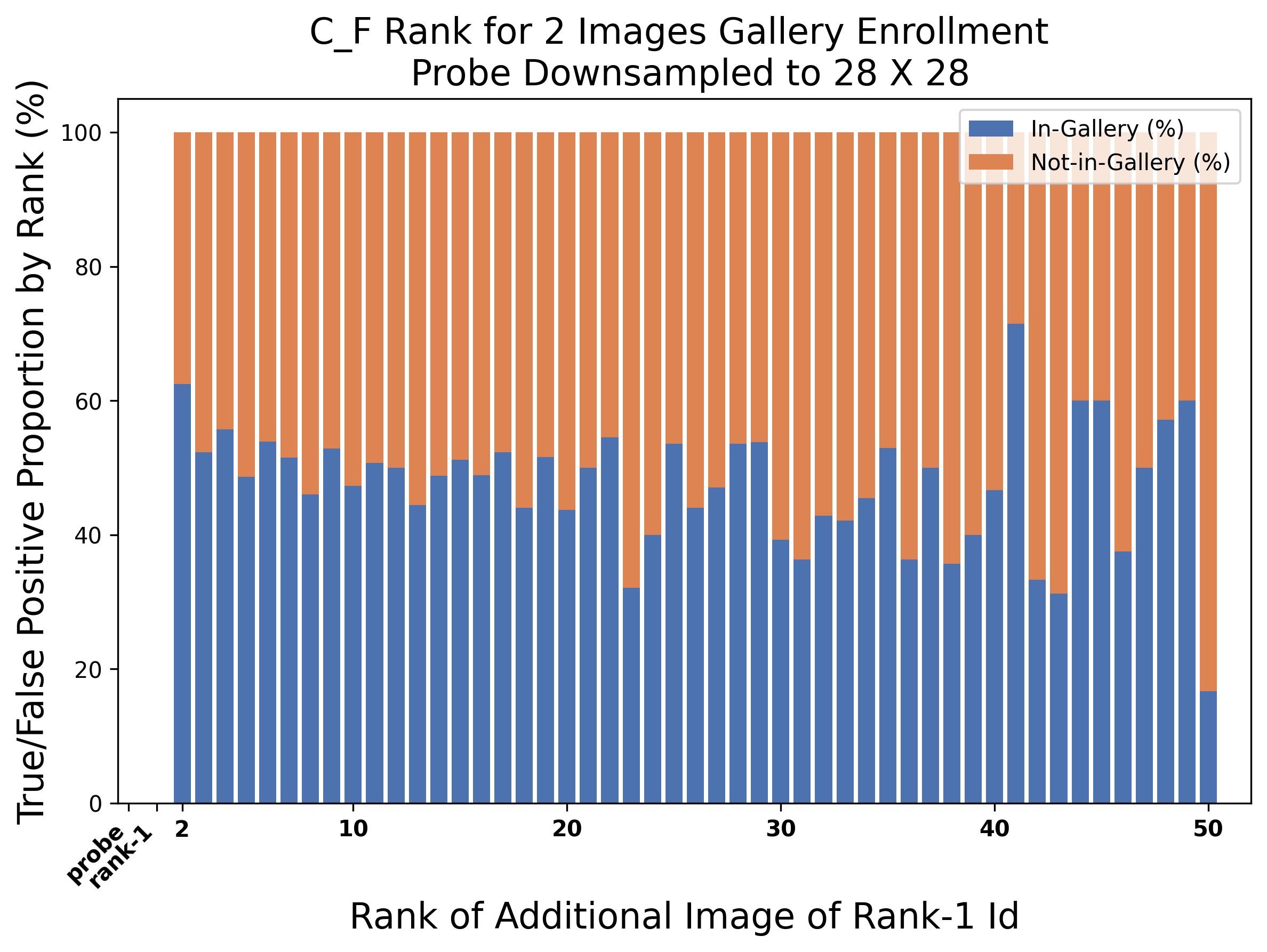}
      \end{subfigure}
      \hfill
      \begin{subfigure}[b]{0.24\linewidth}
        \centering
          \includegraphics[width=1\linewidth]{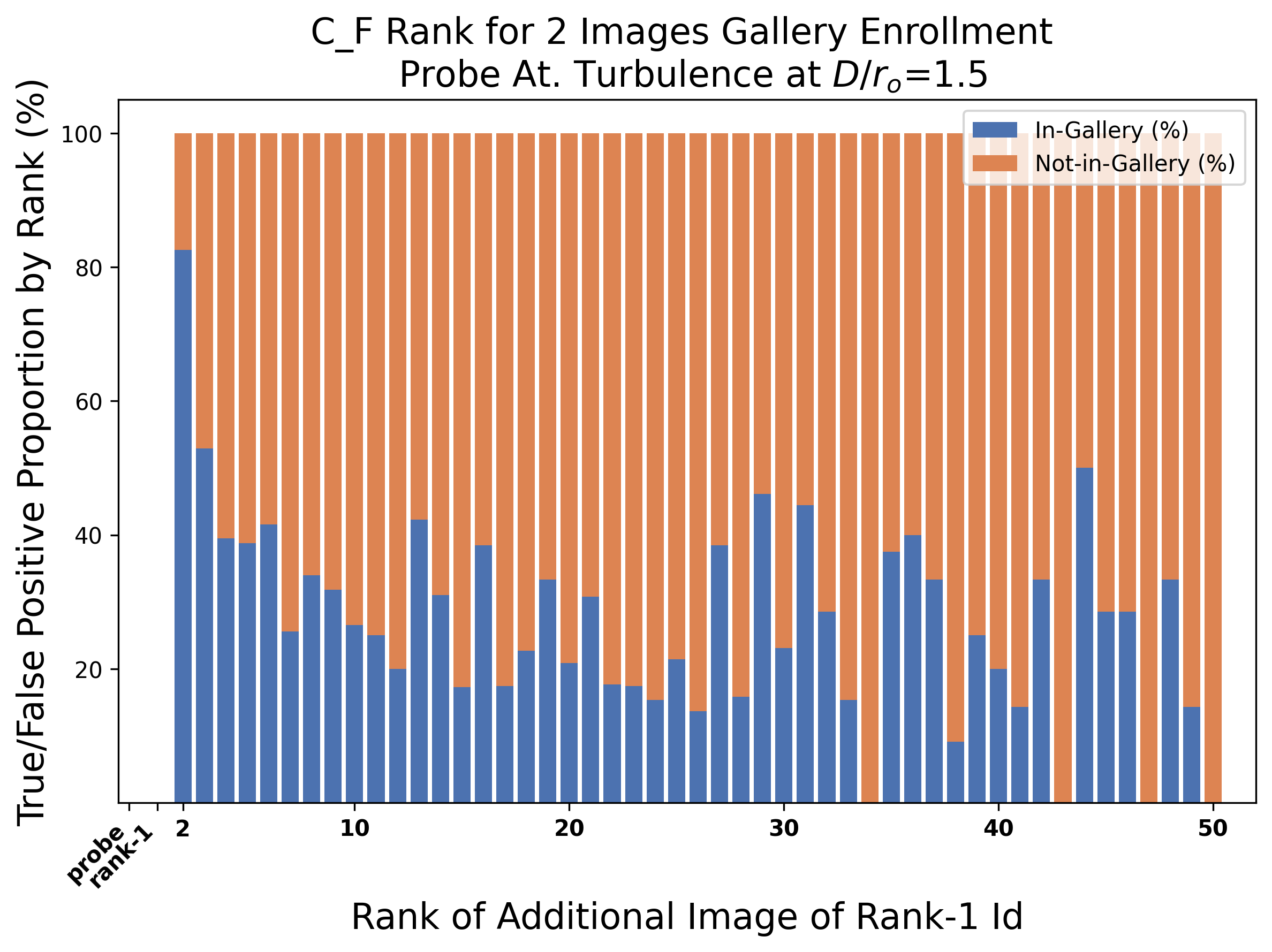}
      \end{subfigure}
       \vspace{0.25em}
  \end{subfigure}
  \hfill
  \begin{subfigure}[b]{1\linewidth}
    \centering
      \begin{subfigure}[b]{0.24\linewidth}
        \centering
          \includegraphics[width=1\linewidth]{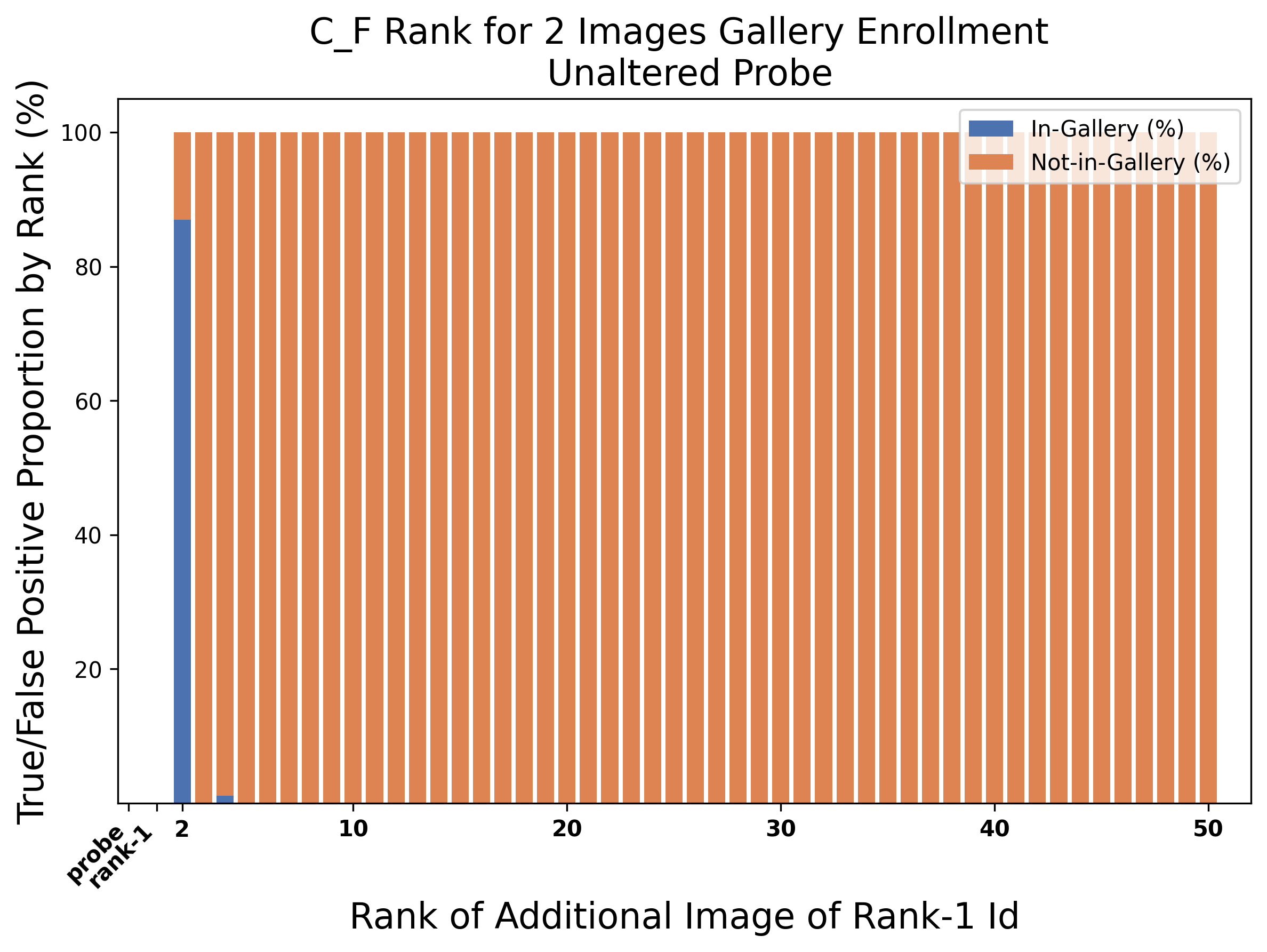}
          \caption{Original Probe}\label{mugshot}
      \end{subfigure}
      \hfill
      \begin{subfigure}[b]{0.24\linewidth}
        \centering
          \includegraphics[width=1\linewidth]{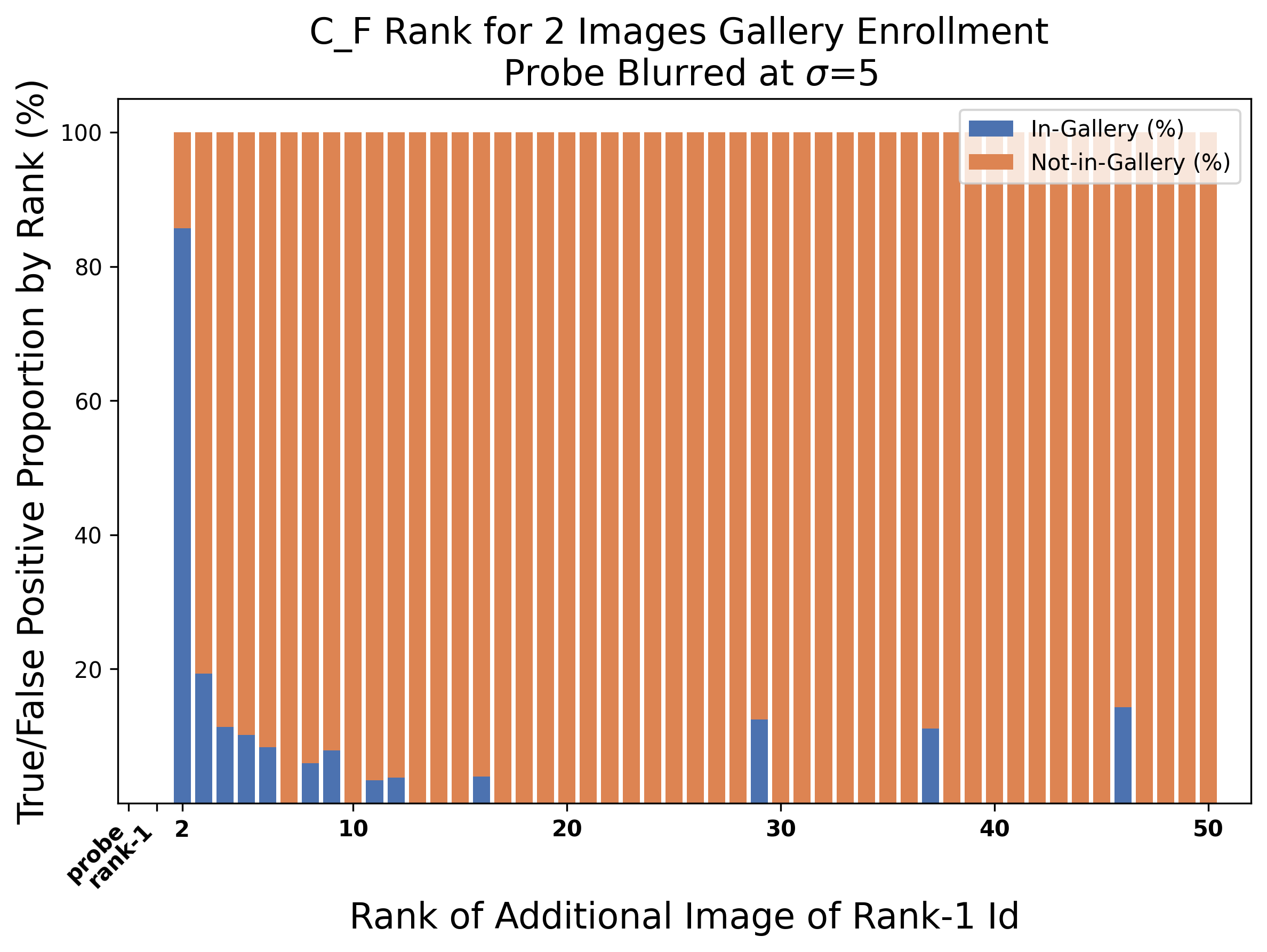}
          \caption{Blurred Probe}
      \end{subfigure}
      \hfill
      \begin{subfigure}[b]{0.24\linewidth}
        \centering
          \includegraphics[width=1\linewidth]{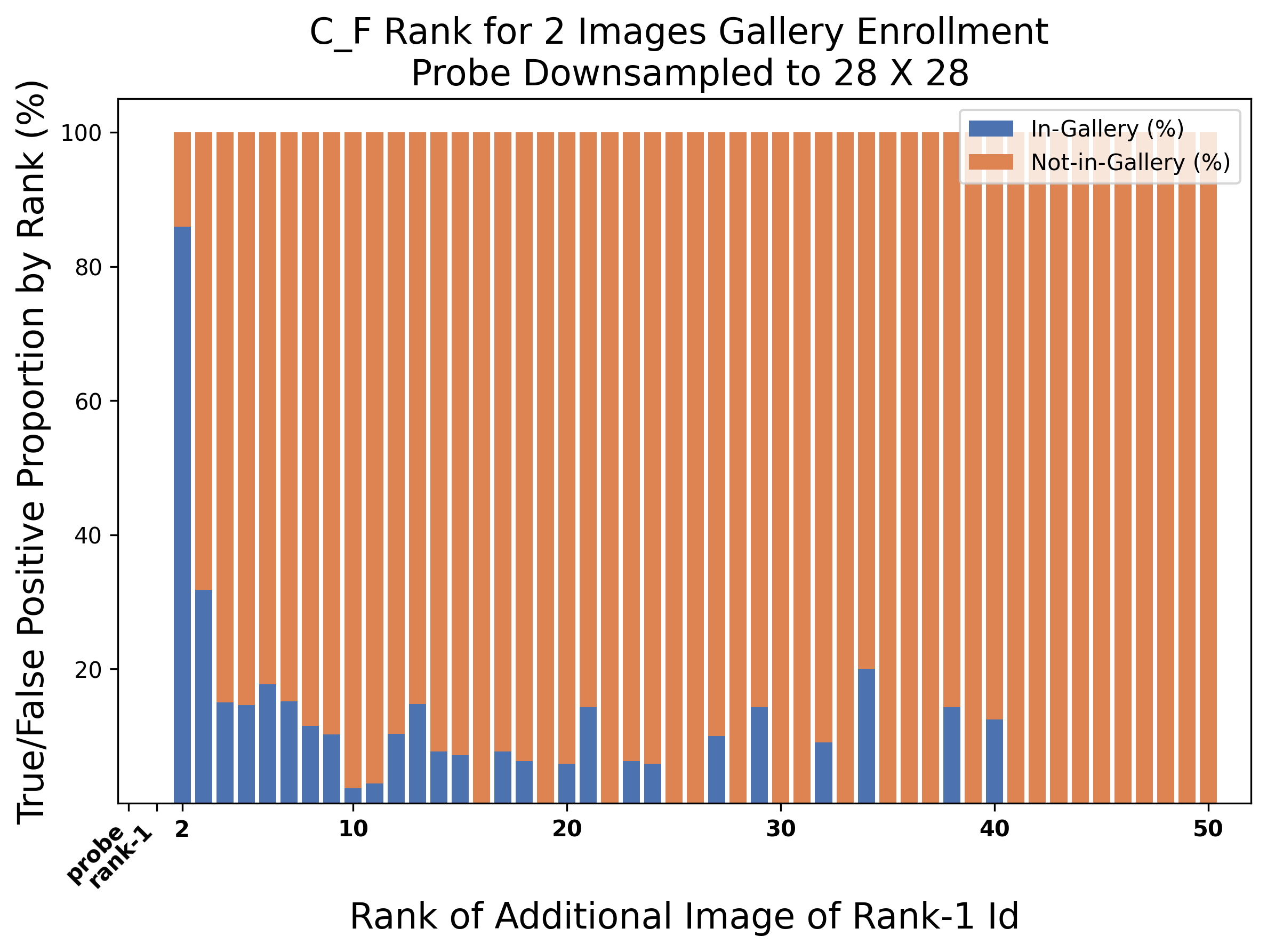}
          \caption{Downsampled Probe}
      \end{subfigure}
      \hfill
      \begin{subfigure}[b]{0.24\linewidth}
        \centering
          \includegraphics[width=1\linewidth]{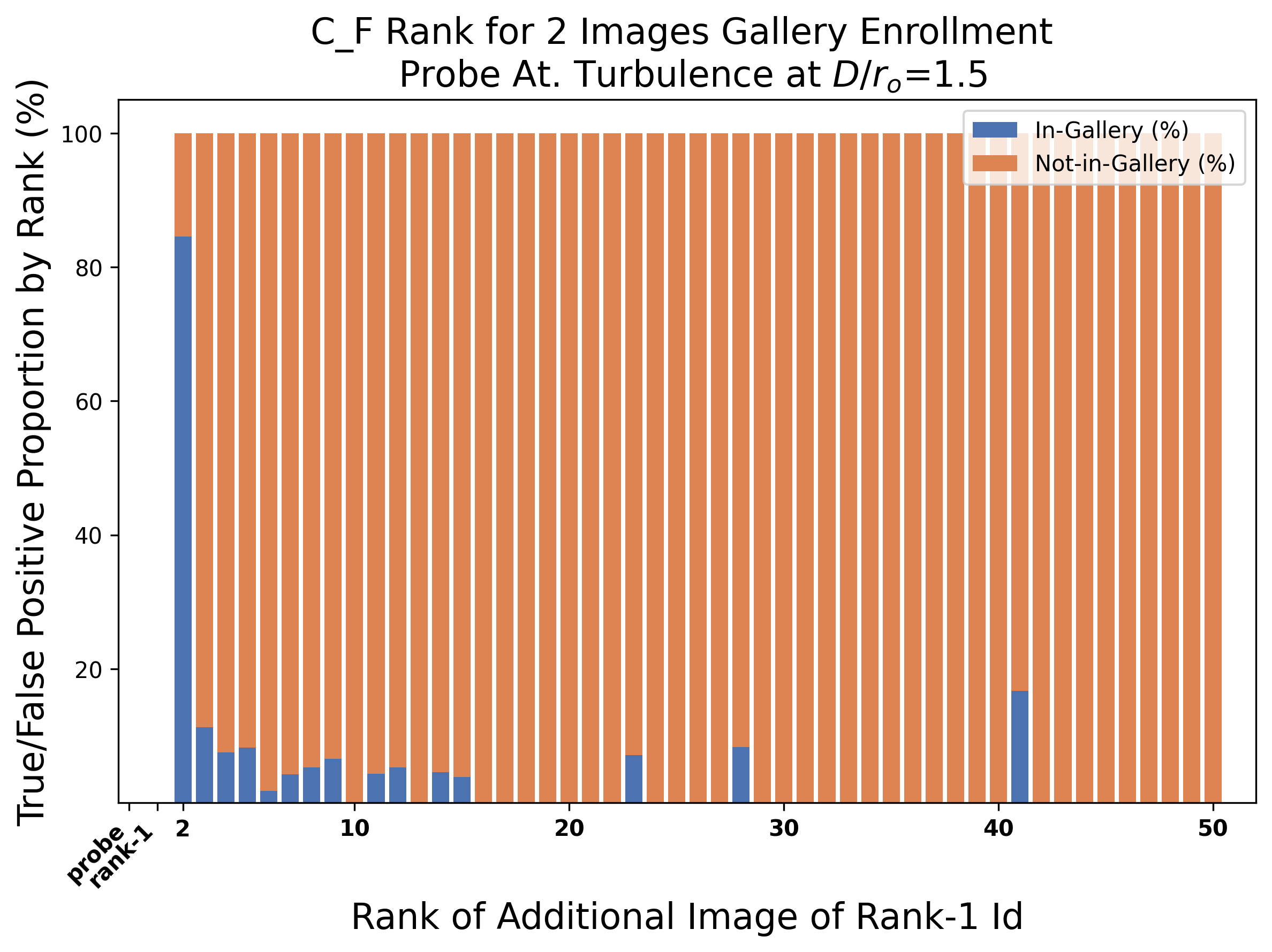}
          \caption{Distorted Probe}
      \end{subfigure}
      \vspace{-0.75em}
  \end{subfigure}
  \caption{ {\bf Rank Distribution of Additional Images for rank-one Identity: In-Gallery vs. Not-In-Gallery Across Probe Qualities and Matchers.} In-gallery searches exhibit more consistent ranking behavior, while not-in-gallery searches show greater variability. Only ranks up to 50 are displayed. The top row presents results using FaceNet, while the bottom row shows results using ArcFace for MORPH Caucasian Females. }
  \vspace{-2em}
  \label{fig:preliminary_results}
\end{figure*}

\noindent{\bf Preliminary Observations.} To examine potential patterns in the rank behavior of the additional images for the rank-one identity, we select all identities with at least three images from  MORPH. The most recent image of each identity serves as the probe, while two images are randomly sampled from the remaining set for enrollment. A 1-to-many search is conducted for both \textit{In-gallery} and \textit{Out-of-gallery} scenarios for each probe, with the results presented in Figure \ref{fig:preliminary_results}. The results in Figure \ref{fig:preliminary_results} focus on the Caucasian female group, as its smaller dataset size makes pattern observation easier. We use two representative matchers: AdaFace, a deep CNN \cite{he_cvpr_2016deep} with a state-of-the-art margin-based loss function, and FaceNet, which represents face recognition networks prior to ResNet and margin-based loss functions. 

Figure~\ref{fig:preliminary_results} shows that when the probe is an \textit{In-gallery} sample, the additional image of the rank-one identity tends to appear at a lower rank compared to when the probe is \textit{Out-of-gallery}. Formally, the conditional probability \( P(in\_gallery \mid r) \) is higher for lower \( r \), indicating a stronger ranking consistency for true positives. For instance, when the probe is a mugshot-quality image (Figure \ref{mugshot}), the probability of an additional image at rank 2 being from an In-gallery search is approximately 85\%. More generally, \( P(in\_gallery \mid r \leq k) \) remains high for small \( k \), confirming that additional ranks of \textit{In-gallery} probe are concentrated in the top ranks. This trend holds even as probe quality degrades for SoTA face matchers, as shown in the bottom row of Figure \ref{fig:preliminary_results}. However, for pre-ResNet and pre-margin-based matchers like FaceNet, the trend of \( P(in\_gallery \mid r \leq k) \) remaining high for small \( k \) no longer holds. This suggests that earlier matchers lack the robustness of modern face embedding methods, exhibiting significant performance degradation for downstream classification task as probe quality decreases—a phenomenon analyzed in detail in Section~\ref{results}.
\vspace{-1em}
\section{Proposed Method}\label{method}

As observed in Section \ref{prelim}, the rank of an additional image producing a rank-one match behaves differently depending on whether the probe is an \textit{In-gallery} or \textit{Out-of-gallery} sample. Modern face recognition systems often operate on multiple images per identity, as noted in the NIST report \cite{grother_nistir_2019b}. While prior work focuses on fusion techniques to enhance match scores or assumes an underlying match score distribution for classification based on extreme value theory, we instead leverage the rank position of additional images to distinguish \textit{In-gallery} from \textit{Out-of-gallery} probes. Since real-world rank patterns are typically non-linear, statistical or linear classifiers may simply learn a trivial and non-reliable threshold-based decision boundary, similar to existing methods. To address this, we train a feedforward neural network to classify \textit{In-gallery} and \textit{Out-of-gallery} samples based on rank patterns of additional images of the rank-one retrieved from a 1-to-many search. More formally, let $\mathbf{x} \in \mathbb{R}^{d_{\text{in}}}$ be the input feature vector. The model applies a sequence of non-linear transformations:
\begin{equation}
    \mathbf{y} = f_{\theta}(\mathbf{x}) = \mathbf{W}_n \cdot g_{n-1}(g_{n-2}(\cdots g_1(\mathbf{x}))) + \mathbf{b}_n
\end{equation}
\noindent where $g_i$ are non-linear mappings:
\begin{equation}
    g_i(\mathbf{h}) = \mathcal{D}(\sigma(\mathcal{N}(\mathbf{W}_i \mathbf{h} + \mathbf{b}_i))), \quad i \in \{1,2,\ldots,n-1\}
\end{equation}
\noindent with:
\begin{itemize}
    \item $\mathcal{N}(\cdot)$ = normalization (e.g., LayerNorm)
    \item $\sigma(\cdot)$ = activation (e.g., ReLU)
    \item $\mathcal{D}(\cdot)$ = dropout ($p$) \\
\end{itemize}

\begin{figure*}[!htbp]
    \captionsetup[subfigure]{labelformat=empty}
    \raggedright
    \centering
    \includegraphics[width=.95\linewidth]{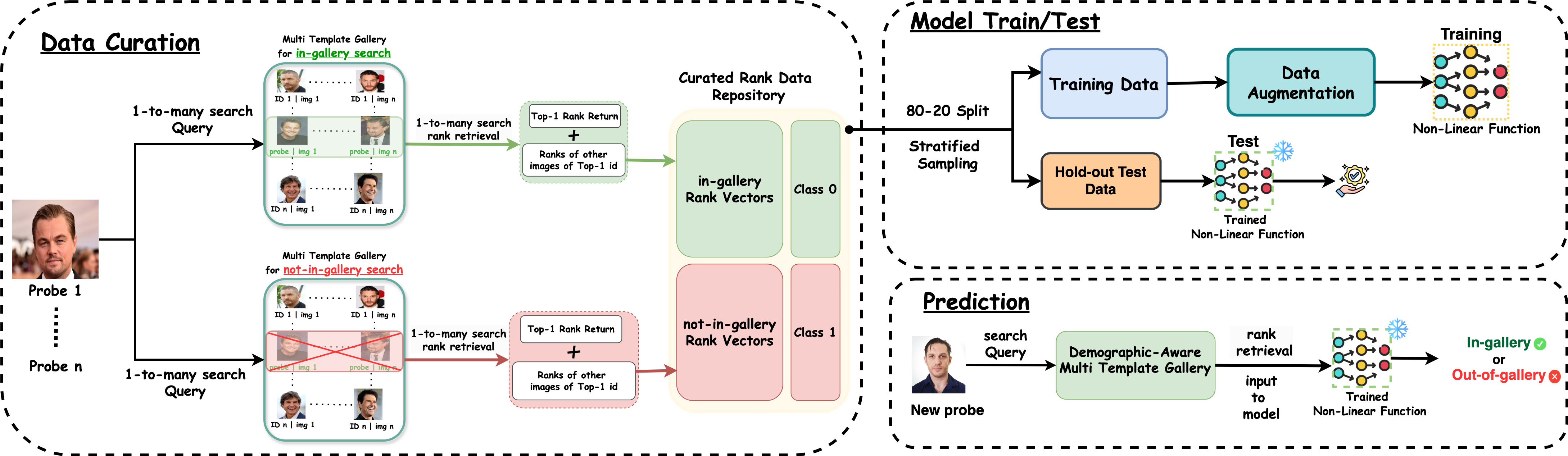}
    \caption{{\bf Overview of the Proposed In-Gallery vs. Not-In-Gallery Classification Method.} Each probe undergoes two 1-to-many searches: (1) \textit{in-gallery}, where probe images are present, and (2) \textit{not-in-gallery}, where they are absent. Each search returns rank vectors of additional images for the rank-one identity, which are then used to train a non-linear classification function to learn this pattern.
}
    \label{fig:Overall}
    \vspace{-1.75em}
\end{figure*}
\noindent The trainable parameters are $\theta = \{\mathbf{W}_i\}_{i=1}^n \cup \{\mathbf{b}_i\}_{i=1}^n$.

\begin{figure}[!htpb]
    \captionsetup[subfigure]{labelformat=empty}
    \raggedright
    \centering
    \includegraphics[width=.75\linewidth]{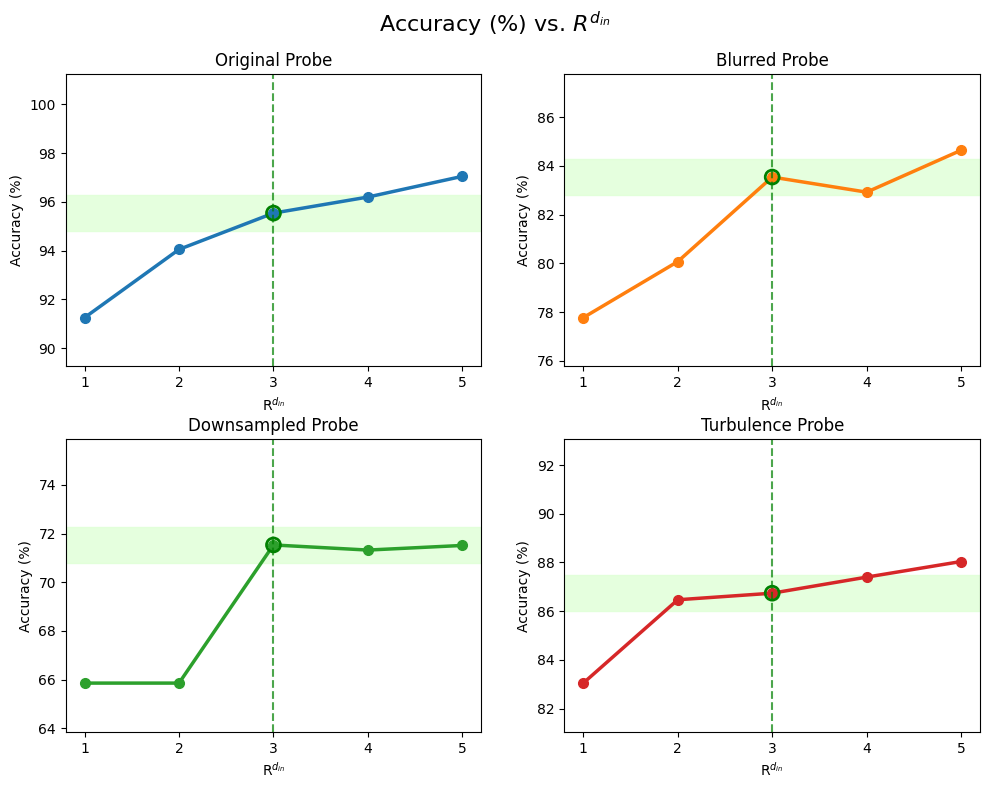}
    \caption{
 Effective $ R^{d_{in}} $ for \textit{in-gallery} vs \textit{not-in-gallery} search.
    }
    \label{fig:cardinality}
    \vspace{-2em}
\end{figure}
\noindent{\bf Rank Cardinality for Effective Learning.} A key factor in training a non-linear model for this binary classification task is determining an appropriate rank vector dimension, \(\mathbb{R}^{d_{\text{in}}}\), needed for effective learning. A one-dimensional rank vector may reduce the model to a simple threshold-based decision, limiting its ability to capture complex patterns and restricting generalization. Increasing the rank vector dimension, e.g., to 10, could enhance accuracy and generalization by capturing richer features. However, this would require at least 12 images per identity, which is often impractical.

Since the model outputs a two-dimensional decision, the input must have at least three dimensions to maintain representational capacity, prevent trivial mappings, and ensure meaningful non-linear decision boundaries while balancing complexity and data constraints. Experimentally, as shown in Figure \ref{fig:cardinality}, classifier performance starts to plateau at \(\mathbb{R}^{d_{\text{in}}} = 3\), indicating that this dimensionality is sufficient for our use case. 
Therefore, we set \(\mathbb{R}^{d_{\text{in}}} = 3\) in all experiments, pairing each probe with four gallery images—the rank-one image and the ranks of three additional images—as input for model training and evaluation.

\noindent{\bf Train-Test Rank Data Curation.} Several open-set (\textit{In-gallery} vs. \textit{Out-of-gallery}) protocols exist, but many have limitations and do not directly apply to our framework, which extends beyond overall classification to demographic-specific probe analysis. The widely used IJB-A dataset provides an open-set protocol but suffers from missing annotations, numerous profile and low-quality images, and large template sizes for both enrollment and querying \cite{gunther_cvprw_2017,klare_cvprw_2015}. Other protocols \cite{ekenel_iccvs_2009,best_tifs_2014,li_tpami_2005,liao_ijcb_2014,stallkamp_iccv_2007} were developed before margin-based loss functions and primarily focus on statistical estimation. In contrast, our framework leverages input ranks for a non-linear classification function. To support this, we introduce an evaluation protocol that aligns with real-world 1-to-many search.

To ensure a robust evaluation, we curate rank data for both \textit{In-gallery} and \textit{Out-of-gallery} samples across different probe quality scenarios, following the preprocessing and selection protocol in Section~\ref{prelim} and aligning with the \(\mathbb{R}^{d_{\text{in}}}\) experiment in Figure \ref{fig:cardinality}. To generate \(\mathbb{R}^{d_{\text{in}}} = 3\), we sample subsets from MORPH and ND-MFAD, selecting identities with at least five images. The most recent image serves as the probe, while the remaining four are randomly sampled and enrolled in the gallery. For MORPH, the number of probe identities is approximately 2K for Caucasian Males (CM), 1K for Caucasian Females (CF), 4.5K for African-American Males (AAM), and 1.5K for African-American Females (AAF). For MFAD, there are approximately 0.5K probe identities for both CM and CF.

To simulate the {\it In-gallery} search, each probe identity is enrolled in the gallery with four additional images. We then perform a 1-to-many search, retrieving the rank-one identity along with the ranks of three other images belonging to this identity. This rank query vector is labeled as {\it In-gallery} (binary label 1). Given the high accuracy of modern deep embeddings, the rank-one identity is almost always the correct match.  For the {\it Out-of-gallery} scenario, we exclude all images of the probe identity from the gallery and repeat the 1-to-many search. Since the probe identity is absent, the rank-one retrieved identity is inevitably a false positive. The ranks of three images belonging to this incorrect rank-one identity form the {\it Out-of-gallery} query vector (binary label 0). This dataset curation and evaluation protocol is simple, intuitive, and ensures a balanced representation of {\it In-gallery} and {\it Out-of-gallery} cases. This curated dataset is split into \textit{disjoint 80-20 training and testing sets}, with \textit{stratified sampling} to maintain class balance.

\begin{table*}[!htpb]
    \centering
    \resizebox{0.9\textwidth}{!}{%
        \subfloat[FaceNet]{
            \begin{tabular}{ccccc}
                \toprule
                Group & Original & Blurred & Downsampled & At. Turbulence  \\
                \midrule
                AA M  & 93.51     & 69.97    & \textcolor{darkred}{54.07}         &  80.09                 \\
                AA F  & 89.91     & 65.14    &  \textcolor{darkred}{ 50.00 }       & 75.39                 \\
                C F   & 87.80    & 70.87    & \textcolor{darkred}{50.00 }     &  77.56                  \\
                C M   & 93.20     & 70.27    & \textcolor{darkred}{54.37}        &  79.13                \\
                \bottomrule
            \end{tabular}
        }\qquad\qquad
        \subfloat[ArcFace]{
            \begin{tabular}{ccccc}
                \toprule
                Group & Original & Blurred & Downsampled & At. Turbulence \\
                \midrule
                AA M  & 95.54     & 83.54    & 71.53        & 86.74                 \\
                AA F  & 94.16     & 79.89    & 67.63        & 84.70                  \\
                C F   & 92.13     & 77.89    & 67.76       & 80.71                   \\
                C M   & 95.27     & 82.09    & 74.19        & 83.98                \\
                \bottomrule
            \end{tabular}
        }
    }
    
    \vspace{0.2em}
    
    \resizebox{0.9\textwidth}{!}{%
        \subfloat[AdaFace]{
            \begin{tabular}{ccccc}
                \toprule
                Group & Original & Blurred & Downsampled & At. Turbulence  \\
                \midrule
                AA M  & 98.07    & 97.47    & 95.65       & 97.52                 \\
                AA F  & 98.58     & 95.27    & 92.90        & 95.53                  \\
                C F   & 97.24     & 95.67    & 86.22        & 94.88                   \\
                C M   & 97.82     & 96.45    & 94.42       & 96.84                 \\
                \bottomrule
            \end{tabular}
        }\qquad\qquad
        \subfloat[TransFace]{
            \begin{tabular}{ccccc}
                \toprule
                Group & Original & Blurred & Downsampled & At. Turbulence  \\
                \midrule
                AA M  & 97.74    & 97.43    & 93.29        & 97.48                  \\
                AA F  & 97.63     & 93.95    & 89.59       & 94.21                  \\
                C F   & 98.43     & 94.09    & 88.58        & 90.94                 \\
                C M   & 98.42     & 97.98    & 93.92       & 96.56                   \\
                \bottomrule
            \end{tabular}
        }
    }
    \vspace{-1em}
    \caption{{\bf Results of \textit{In-gallery} vs. \textit{Out-of-gallery} Classification Across Face Embedding Networks and Varying Conditions.} The results show that for high-quality (mugshot) probes, all face embedding networks—from older models like FaceNet to state-of-the-art (SoTA) models like AdaFace—perform well across different demographic groups. However, as probe quality degrades due to blur, downsampling, or atmospheric turbulence, the performance of older embeddings declines significantly, while newer matchers maintain high accuracy. This validates the reliability of the proposed \textit{In-gallery} vs. \textit{Out-of-gallery} classification, even in real-world search scenarios, particularly when using recent quality-adaptive SoTA face embedding networks. For each face embedding network, results are presented for 16 models, one for each of four demographics and four probe quality conditions. Results are reported on the MORPH dataset.
} 
    \label{fig:MORPH}
    \vspace{-0.5em}
\end{table*}


\begin{table*}[!htpb]
    \centering
    \resizebox{0.9\textwidth}{!}{%
        \subfloat[FaceNet]{
            \begin{tabular}{cccccc}
                \toprule
                Group & Original & Blurred & Downsampled & Sunglasses &  At. Turbulence\\
                \midrule
                C F   & 96.34     & 73.78    & \textcolor{darkred}{50.00}       & 72.96            &  83.54      \\
                C M   & 93.81     & 71.90    & \textcolor{darkred}{50.00}        & 74.76            & 82.86       \\
                \bottomrule
            \end{tabular}
        }\qquad\qquad
        \subfloat[ArcFace]{
            \begin{tabular}{cccccc}
                \toprule
                Group & Original & Blurred & Downsampled & Sunglasses  & At. Turbulence  \\
                \midrule
                C F   & 93.37     & 82.32    & 70.12        & 78.57            & 85.37       \\
                C M   & 95.71     & 83.67   & 71.90        & 83.33           & 85.24       \\
                \bottomrule
            \end{tabular}
        }
    }

    \vspace{0.2em}
    
    \resizebox{0.9\textwidth}{!}{%
        \subfloat[AdaFace]{
            \begin{tabular}{cccccc}
                \toprule
                Group & Original & Blurred & Downsampled & Sunglasses  & At. Turbulence \\
                \midrule
                C F   & 95.12     & 95.73    & 93.90       & 95.12           & 92.07      \\
                C M   & 96.67     & 94.29    & 90.04        & 94.76            & 96.02       \\
                \bottomrule
            \end{tabular}
        }\qquad\qquad
        \subfloat[TransFace]{
            \begin{tabular}{cccccc}
                \toprule
                Group & Original & Blurred & Downsampled & Sunglasses & At. Turbulence \\
                \midrule
                C F   & 96.34     & 95.73    & 92.35        & 93.29            & 95.92       \\
                C M   & 97.62     & 95.22    & 91.63       & 96.19           & 96.42       \\
                \bottomrule
            \end{tabular}
        }
    }
    \vspace{-1em}
    \caption{{\bf Results of \textit{In-gallery} vs. \textit{Out-of-gallery} Classification on the MFAD Dataset.} Results are consistent with observations from MORPH, demonstrating that the findings are not  artifacts of a single dataset but are generalizable across different face datasets.  
}
    \label{fig:MFAD}
    \vspace{-1.75em}
\end{table*}

\noindent{\bf Network and Training Detail.} We use this labeled training data to train a classifier to determine whether a rank feature vector originates from an In-gallery or Out-of-gallery probe search.
{\it The exact form of classifier used for this purpose is not an essential point to our concept.} Our non-linear learning function is a multi-layer perceptron (MLP) with two hidden layers, each containing 16 neurons, incorporating Layer Normalization, ReLU activation, and dropout for improved stability and regularization. The output layer consists of two neurons, representing the probability distribution over the two classes, allowing the model to estimate confidence scores for In-gallery and Out-of-gallery classifications. We employ 10-fold cross-validation to partition the data into training and validation subsets, ensuring robust performance evaluation. The model is trained using the Adam optimizer with a learning rate of \(1 \times 10^{-3}\) and CrossEntropyLoss, updated via mini-batch gradient descent with a batch size of 32 over 20 epochs. Validation accuracy is computed across folds, and the best-performing model from cross-validation is selected for final evaluation on the held-out test set. Since rank patterns vary across conditions and groups, we train separate models to capture the underlying patterns for different groups and probe quality conditions. The overall methodology is illustrated in Fig. \ref{fig:Overall}. 

\noindent{\bf Data Augmentation for Robust Classification.} Given \( n \) ordinal rank inputs, the model may learn overly simplistic decision boundaries.
To mitigate this bias, we apply feature permutation augmentation during training, disrupting the strong feature-to-feature correlation typically observed in rank data 
This method randomly permutes feature values within each sample, preserving the overall distribution while introducing beneficial variability. Since labels remain unchanged, the network learns robust representations invariant to feature order. Additionally, the increased training set diversity enhances generalization on test set.

\vspace{-1em}

\section{Experimental Results}\label{results}

We follow the data curation, augmentation, network, and training procedures outlined in Section \ref{method} to obtain the results presented in Tables \ref{fig:MORPH} and \ref{fig:MFAD}. These results are further analyzed by demographic groups, as discussed in Section \ref{cross-demo}.

\noindent{\bf Mug-shot Quality Probe.} Table \ref{fig:MORPH} shows that when the probe remains unaltered (mug-shot quality), the classifier achieves high accuracy across all face embedding networks. Classification accuracy for In-gallery and Out-of-gallery cases ranges from 87.80\% with FaceNet embeddings to 98.58\% with AdaFace embeddings. These trends are consistent with Figure \ref{fig:preliminary_results}, where FaceNet exhibits cases where the second enrolled image of the rank-one identity in an In-gallery probe ranks significantly lower than rank-one, whereas AdaFace maintains a more stable ranking. The results remain consistent across both MORPH and MFAD.

\noindent{\bf Blurred Probe.} As the probe image used for search is blurred (see Figure \ref{fig:sample_images}), the accuracy of In-gallery vs. Out-of-gallery classification decreases. The decline is particularly severe for FaceNet and notably significant for ArcFace. In contrast, AdaFace and TransFace exhibit a reduction in accuracy, but the drop is less pronounced, and their performance remains relatively strong. These trends are consistent across both the MORPH and MFAD datasets.

\noindent{\bf Downsampled Probe.} Among all tested conditions, downsampling has the most significant impact on accuracy. All face embedding networks experience a substantial drop in performance. For FaceNet embeddings, accuracy drops to 50\%, indicating that with a downsampled probe, the rank patterns for In-gallery and Out-of-gallery samples become nearly arbitrary. These results remain consistent across both the MORPH and MFAD datasets.

\noindent{\bf Distorted Probe.} When the probe image is augmented to simulate atmospheric turbulence (see Figure \ref{fig:sample_images}), the accuracy of In-gallery vs. Out-of-gallery classification decreases. The decline is particularly severe for FaceNet and notably significant for ArcFace. In contrast, AdaFace and TransFace experience a reduction in accuracy, but the drop is less pronounced, and their performance remains relatively strong. These trends are consistent across both the MORPH and MFAD datasets.

\noindent{\bf Probe w/ Sunglasses.} This experiment was conducted only on the MFAD dataset, as a sunglasses-wearing image is available as a probe image for each identity. The accuracy decreases across all embedding networks, with a significant drop for FaceNet and ArcFace, while AdaFace and TransFace experience a more moderate decline.
\vspace{-0.5em}

\section{Comparison with Other Methods}
In this section, we compare the accuracy of our method for \textit{In-gallery} vs. \textit{Out-of-gallery} classification against other approaches. All comparisons are conducted on the AA\_M subset of MORPH, which was used to evaluate our method. Since AdaFace outperforms other matchers on our downstream task (Table \ref{fig:MORPH} and \ref{fig:MFAD}), we use it as the face embedding network for all comparisons.

We compare our method against the standard thresholding approach, following the procedure in \cite{bhatta_wacvw_2024} to determine the threshold. Since our method uses four images per probe, we generate mated and non-mated score distributions from the training set, using cosine similarity for the latter. The threshold is set at the score corresponding to a 1-in-10,000 FPIR in closed-set identification.  For each probe search in \textit{In-gallery} and \textit{Out-of-gallery} scenarios, we compute the maximum scores. An In-gallery score below the threshold is a miss, while a Out-of-gallery score above it is a false positive, following \cite{grother_nistir_2019b}. 

As shown in Table \ref{others}, standard thresholding performs significantly worse than our approach. We further compare our method with statistical classifiers, specifically the mean and median classifiers. The \textit{mean classifier} computes the mean vector for each class (0 and 1) during training and assigns a test point to the class with the closest mean based on Euclidean distance. Similarly, the \textit{median classifier} computes the median vector for each class and assigns labels using Euclidean distance, offering greater robustness to outliers than the mean classifier.  As shown in Table \ref{others}, the mean classifier performs poorly, while the median classifier improves upon it but remains less effective than an MLP. This suggests that the rank patterns of additional images for a rank-one identity exhibit a non-linear structure, which may not be accurately captured by linear or first-order statistics-based classifiers.

We also compare our method against a state-of-the-art feature fusion technique, which integrates match scores from multiple images to obtain a higher-quality matching score. In general, feature fusion outperforms approaches that rely solely on the maximum score from multiple images. Performance further improves through quality-aware fusion strategies such as CAFace \cite{kim_neurips_2022}, as indicated by its superior results under degraded probe conditions. Nevertheless, our MLP-based approach still achieves better overall performance across both mugshot-quality and degraded-probe scenarios.
\vspace{-1em}

\setlength\extrarowheight{5pt}
\begin{table}[!htpb]
\centering
\resizebox{1\columnwidth}{!}{%
\begin{tabular}{l|l|cccc}
        Category                  &   Method      & Orig. & Blur & Down. & Turb.\\ \hline
Thresholding & Max. Score                 &  71.35        & 72.47     &  71.58  & 71.64       \\ \hline
Statistical Classifier & Mean Classifier         &  68.65        &  66.45    & 67.22 & 68.15         \\
& Median Classifier                  & 83.22         & 83.28     &  82.23     &  83.88   \\ \hline
Gallery Image Fusion & Naive Feat. Fusion           & 86.72         & 92.62     & 92.90 &  91.53         \\ 
 & CAFace            & 87.12         &  92.77    &\textcolor[HTML]{3339FF}{\bf  93.51} &  91.96        \\ \hline
(n Gallery Rank Classifier) & MLP & \textcolor[HTML]{3339FF}{\bf  97.74 }       & \textcolor[HTML]{3339FF}{\bf  97.43 }    & \textcolor[HTML]{438763}{\bf  93.29 } & \textcolor[HTML]{3339FF}{\bf 97.48 }        \\

\end{tabular}
}
\vspace{-0.5em}
\caption{{\bf Comparison of methods using AdaFace features for the African-American group for In-Gallery vs Out-of-Gallery (\%) comparison}. Our approach, utilizing an \( n \)-ranks classifier, achieves superior overall performance. [Keys: \textcolor[HTML]{3339FF}{\textbf{Best}}, \textcolor[HTML]{438763} {\textbf{Second Best}}]} \label{others}
\vspace{-2em}
\end{table}


\section{Discussions.}\label{discussions}
\vspace{-0.5em}
\noindent{\bf Performance Across different quality probes.} This work presents the first analysis of In-gallery vs. Out-of-gallery classification under varying probe quality conditions, evaluating the impact of real-world degradations on rank-based patterns. Our results show that older embeddings like FaceNet struggle significantly, particularly under downsampling, where accuracy drops to near-random levels. In contrast, modern matchers such as AdaFace and TransFace exhibit greater robustness across all degradations, including blur, atmospheric turbulence, and occlusions. While all networks experience some performance decline, the rank-based patterns leveraged by our approach remain effective, especially with state-of-the-art embeddings. These findings highlight the feasibility of our method even in challenging real-world scenarios.

\noindent{\bf Underlying Non-Linear Learnable Function.} For binary classification of In-gallery vs. Out-of-gallery query vectors, we employ a simple feedforward network, emphasizing the viability of this approach rather than the architecture itself. While a linear boundary may suffice in controlled settings with high-quality probes and a well-matched gallery, real-world deployments rarely meet these conditions. A non-linear decision function is therefore more suitable, with its complexity depending on gallery size and potentially requiring more complex architectures for scalability in large-scale operations to ensure robust performance.

\noindent{\bf Handling Variable Image Enrollment Conditions.} We present experimental results where each gallery identity has the same number of enrolled images. In real-world scenarios, this varies across identities, but our approach naturally generalizes to handle it. We train an In/Out-of-gallery classifier for each enrollment size up to the point where accuracy plateaus. Referring to Figure \ref{fig:cardinality}, this plateau occurs at four enrolled images in our experiments. If the rank-one match has up to this number of enrolled images, the corresponding classifier is used. For identities with more enrolled images, we randomly sample to create a feature vector for the largest In/Out-of-gallery classifier. This process can be repeated N times to leverage all available images, with results averaged. While our approach requires at least one additional enrolled image of rank-one identity for In-gallery/Out-of-gallery classification during inference, it effectively handles variable gallery sizes.
\vspace{-0.5em}

\section{Conclusion.}\label{conclusion}

We have demonstrated the feasibility of determining whether a rank-one match for an unknown probe image belongs to the gallery or is out-of-gallery by leveraging the ranks of additional images associated with the rank-one identity. Our results are consistent across multiple face embedding networks, with this phenomenon emerging most strongly in those trained using advanced margin-based loss functions—an outcome not observed in simpler embedding approaches. Moreover, the proposed method outperforms standard thresholding, first-order classifiers, and multiple-gallery image fusion techniques, further highlighting its effectiveness. Finally, the method remains effective even under challenging conditions such as blur, reduced resolution, and occlusions (e.g., sunglasses), demonstrating its robustness in maintaining reliable In-gallery vs. Out-of-gallery predictions and underscoring its practical applicability in real-world 1-to-many search.

{
    \small
    \bibliographystyle{ieeenat_fullname}
    \bibliography{main}
}

\end{document}